\documentclass[runningheads]{llncs}

\usepackage{graphicx}
\usepackage{xcolor}
\usepackage{amsmath}
\usepackage{amssymb}
\usepackage{booktabs}
\usepackage{multirow}
\usepackage{microtype}
\usepackage{enumitem}
\usepackage{placeins}
\usepackage{float}

\newcommand{\method}{\textsc{AVEX-Prune}}
\newcommand{\avcap}{\textsc{AVCaps}}
\newcommand{\cmark}{\ensuremath{\checkmark}}
\newcommand{\xmark}{\ensuremath{\times}}

\begin{document}

\title{Audio-Visual Exchange-Aware Token Pruning for Efficient Audio-Visual Captioning}

\titlerunning{Audio-Visual Exchange-Aware Token Pruning}
\authorrunning{Z. Meng et al.}
\author{
Zihan Meng,
Dexiang Hong,
Weidong Chen\thanks{Corresponding author.},
Ziyu Zhou,
Bo Hu,
Zhendong Mao
}
\institute{
University of Science and Technology of China \\
\email{\{zh\_meng,hongdexiang,zhouziyu1204\}@mail.ustc.edu.cn} \\
\email{\{chenweidong,hubo,zdmao\}@ustc.edu.cn}
}

\maketitle

\begin{abstract}
Audio-visual captioning generates natural language descriptions from video and audio content. Multimodal LLMs have advanced this task, but both modalities contribute many tokens to the LLM input, where prefill self-attention scales quadratically. Existing token-pruning methods usually retain tokens by attention, saliency, or cross-entropy loss, yet the hard threshold selection makes it difficult to retain tokens that are truly valuable, especially for high-confusing tokens near the decision boundary. 
To this end, we propose a \method{}, an RL-based audio-visual dynamic token pruning method in this work. In our \method{}, an audio-visual token exchange strategy is proposed to select truly valuable tokens by replacing low-confidence retained tokens with high-confidence candidate tokens from the same or the other modality, and measuring the differences in caption generation from token swaps. \method{} preserves full-token quality at a 40\% retention ratio on both VILA~1.5-8B (54.5 vs.\ 54.6) and VideoLLaMA~2 (57.0 vs.\ 56.8).\footnote{Code will be released in the final version of the paper.}
\end{abstract}


\section{Introduction}
\label{sec:intro}

Audio-visual captioning aims to generate natural language descriptions from both the visual and acoustic content of videos. Video large language models (Video-LLMs) with audio-visual capabilities, such as Video-LLaMA~\cite{zhang2023videollama}, VideoLLaMA~2~\cite{cheng2024videollama2}, and VILA~1.5~\cite{lin2024vila}, jointly encode visual and acoustic streams for this task. Both modalities contribute tokens to the LLM input, where prefill self-attention scales quadratically. Token pruning is therefore necessary, especially as longer videos introduce hundreds of visual patch tokens alongside temporally aligned audio tokens.

\begin{figure}[!t]
    \centering
    \includegraphics[width=1.0\textwidth]{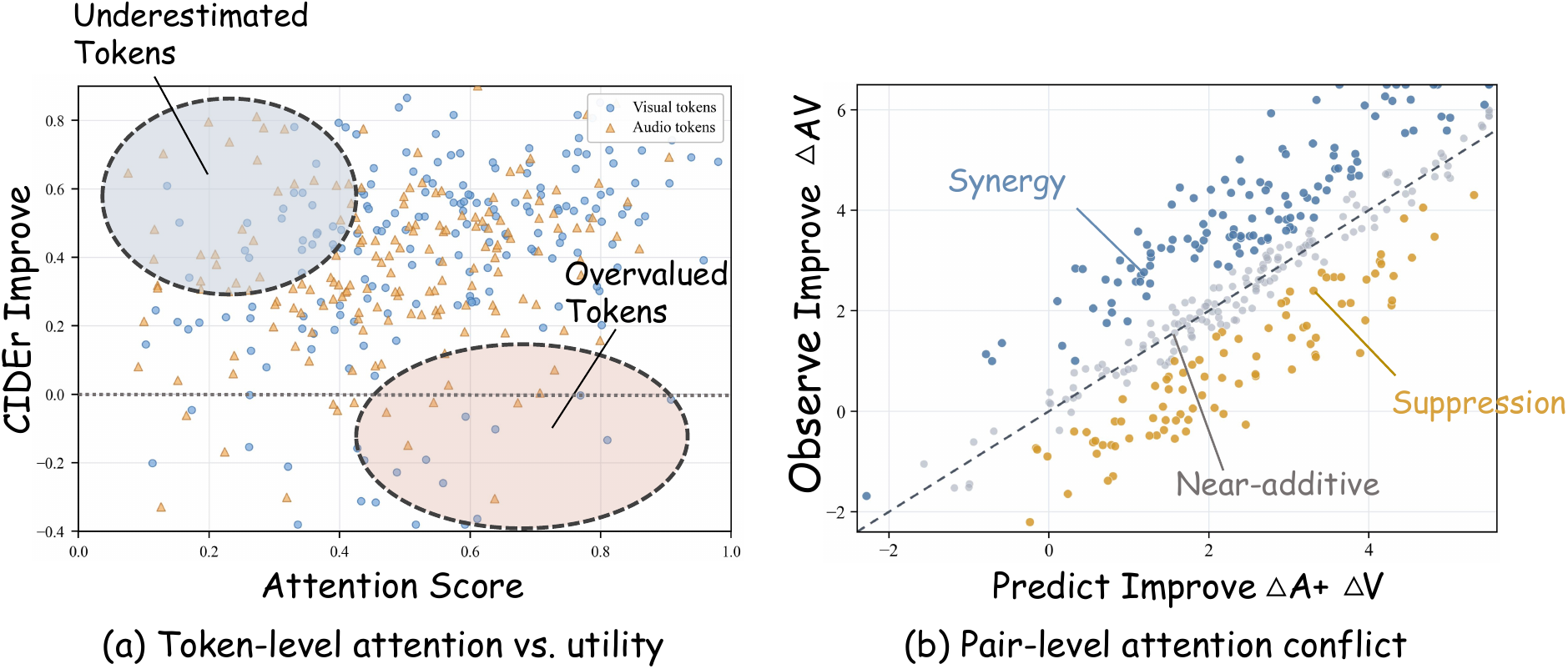}
    \caption{Motivation for exchange-aware audio-visual pruning.
    (a) Counterfactual CIDEr gain vs.\ attention rank: high-attention tokens can contribute
    negligibly, while low-attention tokens may be critical.
    (b) Non-additivity test: the joint CIDEr gain from retaining visual and audio groups
    together deviates from the sum of retaining each group alone.}
    \label{fig:motivation}
\end{figure}

Two challenges motivate our work. First, common pruning proxies (attention weights, visual
saliency, or cross-entropy loss) correlate poorly with token-level contribution to caption
quality. Fig.~\ref{fig:motivation}(a) illustrates this through counterfactual token removal:
many high-attention tokens yield negligible CIDEr gains when retained, while certain
low-attention tokens contribute substantially. Second, audio-visual pruning cannot be reduced
to visual pruning with extra audio tokens. Visual tokens are dense and highly redundant across
frames and spatial regions; audio tokens are sparse, event-driven, and temporally localized
around acoustic events. A global ranking systematically under-selects audio due to the
numerical dominance of visual tokens, while a predetermined audio-visual ratio cannot adapt to diverse
acoustic conditions ranging from silent clips to music-heavy scenes. Fig.~\ref{fig:motivation}(b)
confirms this through an additivity test: the joint gain from retaining both a visual and an
audio group deviates systematically from the sum of retaining each group alone, indicating
that audio-visual contribution is non-additive. These observations indicate that audio and
visual tokens should be ranked through their measured interaction, not as independent pools.

Existing pruning methods address these issues only in part. Proxy-based methods such as
FastV~\cite{chen2024fastv}, SparseVLM~\cite{zhang2024sparsevlm}, and VATP~\cite{guo2024vatp} rank
tokens by attention or saliency without verifying whether highly ranked tokens improve caption
output. Redundancy-reduction methods such as ToMe~\cite{bolya2023tome} and
PruneVid~\cite{huang2025prunevid} merge similar visual tokens, but treat audio-visual pruning as a single pool. FastAV~\cite{jung2026fastav} targets audio-visual pruning with
heuristic scores. CaCoVID~\cite{ma2026cacovid} trains an RL policy for video token pruning, but
its set-level reward leaves a credit-assignment gap: the reward signals which retained set is
better, but cannot attribute the gain to visual evidence, acoustic evidence, or their interaction.

We propose \method{} (Audio-Visual EXchange-Aware Token Pruning), an RL-based dynamic token
pruning method for audio-visual captioning. Instead of relying on attention or saliency proxies,
\method{} learns token selection from actual caption reward improvement. During training, an
audio-visual token exchange strategy replaces low-confidence retained tokens with high-confidence
candidate tokens from the same or the other modality, using four structured replacements:
visual-to-visual, audio-to-audio, visual-to-audio, and audio-to-visual. The frozen captioner
generates captions for these sets, and the CIDEr differences from the token swaps serve as
pairwise supervision for the pruning policy, without learning a value model or updating the LLM.
At inference, all exchange and reward computation is removed, leaving only a single policy forward
pass and Top-$K$ selection. On \avcap, \method{} preserves full-token quality at a 40\% retention
ratio on both VILA~1.5-8B (54.5 vs.\ 54.6 CIDEr) and VideoLLaMA~2 (57.0 vs.\ 56.8 CIDEr).

Our contributions are summarized as follows:
\begin{enumerate}[leftmargin=*, itemsep=0pt, topsep=0pt]
    \item We propose an RL-based framework named \method{} for dynamic token pruning in audio-visual captioning in this work. Instead of relying on attention, saliency, or loss proxies in existing works, our RL reward is trained on the differences in caption generation from token exchanges, which helps select truly valuable tokens from high-confusing tokens near the decision boundary.

    \item In our \method, an audio-visual token exchange strategy is proposed to select truly valuable tokens by replacing low-confidence retained tokens with high-confidence candidate tokens from the same or the other modality. The four structured replacements (visual-to-visual, audio-to-audio, visual-to-audio, and audio-to-visual) allow training to explore better combinations of visual and acoustic evidence.

    \item On \avcap, \method{} preserves full-token quality at a 40\% retention ratio and consistently outperforms pruning baselines on VILA 1.5 and VideoLLaMA~2, with modality-specific gains showing stronger visual-acoustic token combinations.
\end{enumerate}

\section{Related Work}
\label{sec:related}

\subsection{Audio-visual captioning and Video-LLMs.}
Video-LLaMA~\cite{zhang2023videollama}, VideoLLaMA~2~\cite{cheng2024videollama2}, and
VILA~\cite{lin2024vila} connect visual and audio encoders to instruction-tuned LLMs, building
on vision-language pre-training~\cite{radford2021clip,liu2023llava,li2023blip2,alayrac2022flamingo,huang2025graphmoe,lin2024prompting,wang2023structured,han2023textstyle,jin2024multigrained,tian2023aspect,zhou2025hierarchical,wang2025microaction,wang2023contour,liu2025matching,zhao2023difference}
and video-language models~\cite{tang2023vid2seq,li2023videollava,wu2023valley,li2023videochatgpt,lin2023videolora,li2023mplugowl,li2023instructblip,wang2022git,girdhar2023imagebind,chen2021ccanet,chen2022clipmil,chen2022manet,ye2024dualpath,ye2025mrem,chen2026soecgn,ye2025lscan,song2025amdnet,qin2025qcm,li2024visualrel,fu2024sotvae,jin2024d2net,liu2024bootstrapping,li2025pseudo,guo2025emoverse,wang2026merg,zhang2026creatidesign,chen2026creatiparser,zhang2026stimuli,chen2026facenet}.
\avcap{}~\cite{sudarsanam2025avcaps} provides separate visual, audio, and audio-visual captions
per clip, enabling modality-specific evaluation under pruning.

\subsection{Token pruning with retention ratio control.}
Classic token reduction for ViTs~\cite{rao2021dynamicvit,liang2022evit,ryoo2021tokenlearner,song2023vidtome} demonstrated the effectiveness of pruning and merging. FastV~\cite{chen2024fastv},
SparseVLM~\cite{zhang2024sparsevlm}, and VATP~\cite{guo2024vatp} prune via attention or saliency.
ToMe~\cite{bolya2023tome}, PruneVid~\cite{huang2025prunevid}, HoliTom~\cite{shao2025holitom},
and AIM~\cite{zhong2025aim} exploit visual-token redundancy. MADTP~\cite{cao2024madtp} and
MMS-LLaMA~\cite{yeo2025mmsllama} explore modality-aware token reduction. These methods do not
verify whether proxy-ranked tokens improve caption metrics or consider dynamic per-clip
audio-visual retention ratio.

\subsection{Reward-based pruning.}
CaCoVID~\cite{ma2026cacovid} trains an RL policy for token pruning, but its set-level
reward does not distinguish whether gains come from visual evidence, acoustic evidence, or
their interaction. \method{} uses frozen-LLM reward differences between local replacements
to directly supervise token ranking within and across modalities.

\section{Method}
\label{sec:method}

\subsection{Problem Setup}
\label{sec:setup_problem}

The frozen backbone produces visual tokens $\mathbf{H}_v$, audio tokens $\mathbf{H}_a$, and
text prompt tokens $\mathbf{H}_t$ (never pruned). Let $\mathcal{V}$ and $\mathcal{A}$ denote
visual and audio token index sets, and $\mathcal{M}=\mathcal{V}\cup\mathcal{A}$ with
$N=N_v+N_a$. The policy selects $S\subset\mathcal{M}$ with $|S|=K=\mathrm{round}(\rho N)$.
The reward is $R(S)=\mathrm{CIDEr}(y(S),\mathcal{Y}_{av})$~\cite{vedantam2015cider},
where $y(S)$ is the caption from the frozen LLM and $\mathcal{Y}_{av}$ are \avcap{}'s
audio-visual references. Visual-only and audio-only references define $C_v(S)$ and $C_a(S)$
for evaluation only.

\subsection{Text-Conditioned AVEX Policy}
\label{sec:policy}

Visual and audio tokens from the frozen backbone are concatenated and fed into a two-layer
non-causal Transformer encoder:
\[
Z_{av}=\mathrm{AVEnc}([\mathbf{H}_v;\mathbf{H}_a]).
\]
We then apply text-conditioned cross-attention with AV tokens as queries and text tokens as
keys and values, followed by a residual connection:
\[
\hat{Z}_{av}=\mathrm{CrossAttn}(Q=Z_{av},\,K=\mathbf{H}_t,\,V=\mathbf{H}_t),
\]
\[
Z'_{av}=\mathrm{LN}(Z_{av}+\hat{Z}_{av}).
\]
Each token passes through an MLP and a shared AV Scoring Head (LayerNorm, MLP, linear scalar)
to produce a keep score $s_i$. The first Transformer layer is initialized from the first LLM
block of VideoLLaMA~2 with matching dimensions; remaining layers are randomly initialized.
The policy has $\sim$30M trainable parameters; all encoders, projectors, and LLM weights are frozen.

During training, Gumbel-Top-$K$ makes the sampling process differentiable by defining a
stochastic exact-$K$ policy under a target retention ratio. If the sampled order is $o_1,\ldots,o_K$, its Plackett-Luce
log probability is
\[
\log \pi_\theta(S|x)=\sum_{t=1}^{K}\left[
s_{o_t}/\tau-\log\sum_{j\notin\{o_1,\ldots,o_{t-1}\}}\exp(s_j/\tau)\right].
\]
Rewards are black-box; gradients flow through $\nabla_\theta\log\pi_\theta(S|x)$ and
the exchange losses below. The AVEX loss (Section~\ref{sec:avex}) does not depend on
Gumbel sampling, reducing policy-gradient variance and the train-test gap. At inference,
deterministic Top-$K$ replaces Gumbel-Top-$K$ with negligible performance difference.

\subsection{Audio-Visual Exchange Preference Learning}
\label{sec:avex}

\begin{figure}[t]
    \centering
    \includegraphics[width=1.0\textwidth]{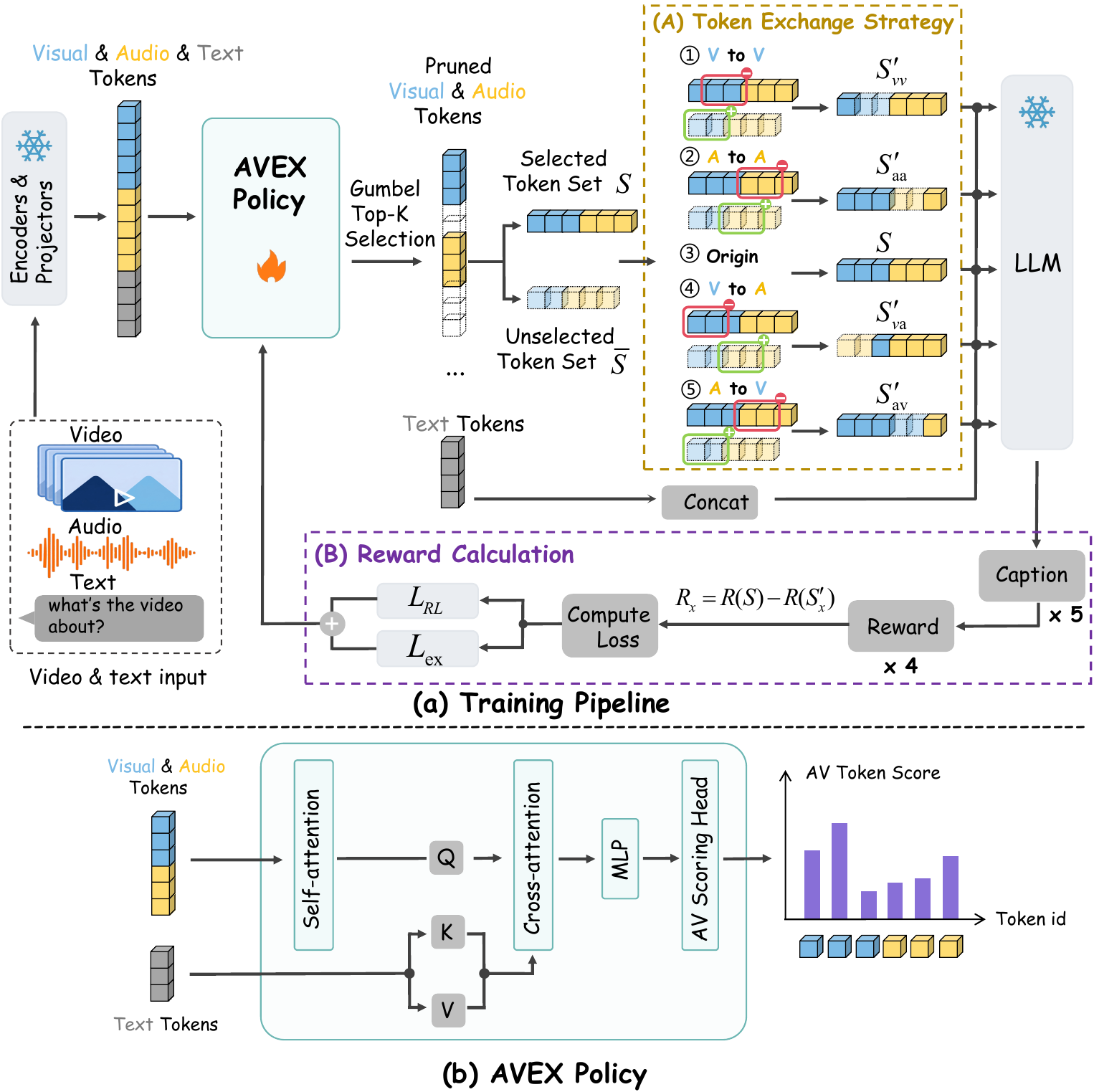}
    \caption{Training framework of \method{}. Four equal-size exchanges compare the sampled
    anchor set with counterfactual sets; CIDEr reward differences supervise group score
    differences and update only the AVEX policy.}
    \label{fig:avex_detail}
\end{figure}

For a sampled anchor set $S$, let $\bar{S}=\mathcal{M}\setminus S$. We construct a
counterfactual set by removing a retained group $G\subset S$ and inserting an equal-sized
candidate group $G'\subset \bar{S}$:
\[
S'=S\setminus G\cup G', \qquad |S'|=|S|=K .
\]
Thus reward differences are caused by evidence replacement while the retention ratio stays constant. The
reward label is $\Delta R=R(S)-R(S')$.

We construct four exchange types: V-to-V ($G\subset S\cap\mathcal{V}$,
$G'\subset\bar{S}\cap\mathcal{V}$), A-to-A ($G\subset S\cap\mathcal{A}$,
$G'\subset\bar{S}\cap\mathcal{A}$), V-to-A ($G\subset S\cap\mathcal{V}$,
$G'\subset\bar{S}\cap\mathcal{A}$), and A-to-V ($G\subset S\cap\mathcal{A}$,
$G'\subset\bar{S}\cap\mathcal{V}$). The first two learn within-modality ranking; the latter
two learn whether audio or visual evidence should replace the other modality at the same
retention ratio.

For reproducibility, groups are selected by score boundaries. Within each exchange, $G$ is
the $g$ lowest-scoring retained tokens from the removed modality, and $G'$ is the $g$
highest-scoring unselected tokens from the inserted modality. We use $g=8$: at a 20\% retention ratio
($K=64$), this is $\sim$12.5\% of the retained set, large enough for stable reward differences but
small enough to avoid conflating many events. If either side has fewer than $g$ tokens, we use
the largest feasible group and skip the exchange if fewer than 2 tokens are available.

The policy prediction for an exchange is:
\[
\Delta F=\sum_{i\in G}s_i-\sum_{j\in G'}s_j .
\]
The AVEX loss is a weighted pairwise preference:
\[
\mathcal{L}_{\mathrm{AVEX}}=
w(\Delta R)\log\left(1+\exp[-\mathrm{sign}(\Delta R)\Delta F]\right),
\quad
w(\Delta R)=\min(|\Delta R|/5,1).
\]
The weight function $w(\Delta R)$ scales linearly up to $|\Delta R|=5$ CIDEr points, then
saturates. If replacing visual tokens with audio improves reward, V-to-A pushes those audio
tokens above the removed visual group. If audio is irrelevant, A-to-V pushes visual candidates
above the removed audio group. Importantly, A-to-V does not uniformly suppress audio tokens;
$w(\Delta R)$ down-weights near-zero exchanges, while $\mathrm{sign}(\Delta R)$ determines
which group should rank higher. 

\subsection{Training Schedule and Objective}
\label{sec:training}

Training has two stages. First, a half-epoch warmup uses cached full-token attention from
text to audio-visual tokens to initialize a balanced policy: $K_a^0=\mathrm{round}(K N_a/N)$
audio and $K-K_a^0$ visual tokens are selected by attention within each modality. Second,
we train with black-box reward optimization and AVEX preference learning. The policy-gradient
term is
\[
\mathcal{L}_{\mathrm{RL}}=-(R(S)-b)\log\pi_\theta(S|x),
\]
where $b$ is the mean reward across anchor and counterfactual sets (with stop-gradient).
The exchange term is
\[
\mathcal{L}_{\mathrm{ex}}=\frac{1}{|\mathcal{E}|}\sum_{x\in\mathcal{E}}\mathcal{L}_{x},
\]
where $\mathcal{E}$ is the set of valid exchanges among V-to-V, A-to-A, V-to-A, and A-to-V.
The final objective is $\mathcal{L}=\mathcal{L}_{\mathrm{RL}}+\lambda_{\mathrm{ex}}\mathcal{L}_{\mathrm{ex}}$
with $\lambda_{\mathrm{ex}}=1$. We train for three epochs with AdamW, learning rate $2\times10^{-5}$,
batch size 64, weight decay 0.01, 5\% warmup, cosine learning rate schedule, and gradient
clipping at 1.0. Reward decoding uses beam size 3, temperature 0, and max length 64.

\section{Experiments}
\label{sec:exp}

\subsection{Experimental Setup}
\label{sec:exp_setup}

\paragraph{Datasets.}
\avcap{}~\cite{sudarsanam2025avcaps} provides 5,125 clips, each with five visual, audio,
and audio-visual caption references. MSRVTT~\cite{xu2016msrvtt} contains 10,000 clips; we
pair YouTube-downloaded audio (7,310 clips with valid audio) with 16-frame visual streams.
We follow the official AVCaps split and use MSRVTT only to evaluate transfer without
training on MSRVTT.

\paragraph{Backbone and baselines.}
Table~\ref{tab:avcaps_base} reports VILA~1.5-8B~\cite{lin2024vila} as an additional reference.
All pruning experiments use VideoLLaMA~2~\cite{cheng2024videollama2} with
Qwen2-7B~\cite{yang2024qwen2}, SigLIP~\cite{zhai2023siglip}, and BEATs~\cite{chen2023beats} as
the frozen backbone, fine-tuned on AVCaps. Each example uses 16 frames and 16\,kHz audio
($N_v=256$, $N_a=64$). Baselines: random pruning, AV extensions of FastV~\cite{chen2024fastv} and
ToMe~\cite{bolya2023tome} (heuristic), FastAV~\cite{jung2026fastav} (heuristic), and an
AV extension of CaCoVID~\cite{ma2026cacovid} (RL-trained). All methods share identical prompt,
decoding (beam 3, temperature 0, max length 64), and retention ratio. Metrics:
modality-specific CIDEr ($C_{av}$, $C_v$, $C_a$), MSRVTT CIDEr, latency, FLOPs, and memory.
Random seed 42. Training: 4$\times$A800, per-GPU micro-batch 1, 16 accumulation steps (effective batch 64),
28\,GB/GPU; only the 30M policy is updated.

\subsection{Captioning Backbone and Reference Models}
\label{sec:base_captioning}

We first establish full-token captioning performance in Table~\ref{tab:avcaps_base}. All
models are fine-tuned on AVCaps; later pruning experiments use VideoLLaMA~2 as the frozen
backbone.

\begin{table}[t!]
    \centering
    \scriptsize
    \setlength{\tabcolsep}{2.0pt}
    \renewcommand{\arraystretch}{1.00}
    \caption{Captioning performance on AVCaps and MSRVTT. VILA~1.5-8B is an additional
    reference; later pruning comparisons use VideoLLaMA~2.}
    \label{tab:avcaps_base}
    \resizebox{\textwidth}{!}{%
    \begin{tabular}{lcccccc}
        \toprule
        \multirow{2}{*}{\textbf{Model}} & \multicolumn{3}{c}{\textbf{AVCaps}} & \multicolumn{3}{c}{\textbf{MSRVTT}} \\
        \cmidrule(lr){2-4}\cmidrule(lr){5-7}
        & \textbf{C} & \textbf{B-4} & \textbf{M} & \textbf{C} & \textbf{B-4} & \textbf{M} \\
        \midrule
        OneLLM (7B)~\cite{han2024onellm} & 27.7 & 11.8 & 26.1 & 52.8 & 37.9 & 18.7 \\
        AV-LLM (13B)~\cite{shu2023avllm} & 30.8 & 13.1 & 29.0 & 55.1 & 39.0 & 21.3 \\
        PandaGPT (13B)~\cite{su2023pandagpt} & 32.0 & 13.6 & 30.1 & 59.4 & 42.0 & 23.9 \\
        Macaw-LLM (7B)~\cite{lyu2023macawllm} & 37.9 & 16.1 & 35.8 & 63.0 & 47.2 & 27.9 \\
        Video-LLaMA (7B)~\cite{zhang2023videollama} & 48.5 & 19.5 & 40.2 & 71.6 & 53.3 & 32.9 \\
        LLaVA-NeXT-Video (7B)~\cite{liu2024llavanext} & 52.3 & 21.3 & 48.5 & 78.0 & 56.3 & 33.1 \\
        OmAgent (7B)~\cite{zhang2024omagent} & 49.5 & 19.8 & 42.2 & 74.1 & 52.9 & 31.5 \\
        \midrule
         VILA~1.5 (8B)~\cite{lin2024vila} & \textbf{54.6} & \underline{22.6} & \underline{47.1} & \underline{79.9} & \underline{56.8} & \underline{33.7} \\
        \textbf{ + \method{} (retention ratio=40\%)} & \underline{54.5} & \textbf{22.8} & \textbf{47.1} & \textbf{80.0} & \textbf{57.2} & \textbf{34.0} \\
        \midrule
        VideoLLaMA~2 (7B)~\cite{cheng2024videollama2} & \underline{56.8} & \underline{23.6} & \underline{49.6} & \underline{80.5} & \underline{57.9} & \underline{35.0} \\
        \textbf{ + \method{} (retention ratio=40\%)} & \textbf{57.0} & \textbf{23.7} & \textbf{49.7} & \textbf{80.7} & \textbf{58.1} & \textbf{35.1} \\
    
        \bottomrule
    \end{tabular}%
    }
\end{table}

\subsection{Main Pruning Comparison on AVCaps}
\label{sec:main}

We next compare pruning methods under the same retention ratio, prompt, decoding, and frozen
VideoLLaMA~2 backbone. Table~\ref{tab:main_pruning} reports modality-specific CIDEr
($C_{av}$, $C_v$, $C_a$) and MSRVTT CIDEr from 50\% to 10\% retention ratio. Rel.~is defined as
$\text{Rel.}=100\times\frac{1}{4}\big(C_{av}/56.8+C_v/52.1+C_a/51.9+C_{\text{MSRVTT}}/80.5\big)$,
averaging the four metrics against the full-token reference. Values moderately above 100
occur when pruning removes noise tokens; Rel.~within $\pm$1 to 2 points of 100 is typical
CIDEr fluctuation.

\begin{table}[t!]
    \centering
    \scriptsize
    \setlength{\tabcolsep}{1.8pt}
    \renewcommand{\arraystretch}{0.82}
    \caption{Pruning comparison on AVCaps and MSRVTT.}
    \label{tab:main_pruning}
    \resizebox{0.85\textwidth}{!}{%
    \begin{tabular}{lcccccc}
        \toprule
        \multirow{2}{*}{\textbf{Method}} & \multirow{2}{*}{\textbf{Retention Ratio}} &
        \multicolumn{3}{c}{\textbf{AVCaps}} & \multicolumn{1}{c}{\textbf{MSRVTT}} &
        \multirow{2}{*}{\textbf{Rel.}} \\
        \cmidrule(lr){3-5}\cmidrule(lr){6-6}
        & & \textbf{$C_{av}$} & \textbf{$C_v$} & \textbf{$C_a$} & \textbf{C} & \\
        \midrule
        Full tokens & 100\% & 56.8 & 52.1 & 51.9 & 80.5 & 100.0 \\
        \midrule
        Random & 50\% & 50.8 & 46.8 & 46.4 & 72.2 & 89.6 \\
        FastV (AV ext.) & 50\% & 56.0 & 51.6 & 51.2 & 79.6 & 98.8 \\
        ToMe (AV ext.) & 50\% & 56.4 & 51.6 & 51.6 & 79.8 & 99.2 \\
        FastAV & 50\% & 56.6 & 52.1 & 51.8 & 80.3 & 99.8 \\
        CaCoVID (AV ext.) & 50\% & 57.1 & 52.2 & 52.0 & 80.8 & 100.3 \\
        \textbf{\method} & 50\% & \textbf{57.6} & \textbf{52.8} & \textbf{52.4} & \textbf{81.6} & \textbf{101.2} \\
        \midrule
        Random & 40\% & 46.5 & 42.9 & 42.5 & 66.0 & 82.1 \\
        FastV (AV ext.) & 40\% & 54.1 & 49.4 & 49.5 & 76.7 & 95.1 \\
        ToMe (AV ext.) & 40\% & 54.7 & 50.3 & 49.9 & 77.5 & 96.4 \\
        FastAV & 40\% & 55.6 & 50.8 & 50.8 & 78.7 & 97.7 \\
        CaCoVID (AV ext.) & 40\% & 56.4 & 51.9 & 51.5 & 80.2 & 99.5 \\
        \textbf{\method} & 40\% & \textbf{57.0} & \textbf{52.3} & \textbf{52.1} & \textbf{80.7} & \textbf{100.1} \\
        \midrule
        Random & 30\% & 41.7 & 38.4 & 38.1 & 59.3 & 73.6 \\
        FastV (AV ext.) & 30\% & 53.0 & 48.4 & 48.4 & 74.8 & 93.1 \\
        ToMe (AV ext.) & 30\% & 53.3 & 48.9 & 48.5 & 75.5 & 93.7 \\
        FastAV & 30\% & 55.2 & 50.4 & 50.4 & 78.0 & 97.0 \\
        CaCoVID (AV ext.) & 30\% & 56.0 & 51.5 & 51.1 & 79.6 & 98.7 \\
        \textbf{\method} & 30\% & \textbf{56.3} & \textbf{51.6} & \textbf{51.4} & \textbf{79.7} & \textbf{98.9} \\
        \midrule
        Random & 20\% & 35.0 & 32.3 & 32.0 & 49.8 & 61.8 \\
        FastV (AV ext.) & 20\% & 51.1 & 47.0 & 46.7 & 72.6 & 90.1 \\
        ToMe (AV ext.) & 20\% & 52.1 & 47.6 & 47.6 & 73.6 & 91.5 \\
        FastAV & 20\% & 53.9 & 49.6 & 49.2 & 76.6 & 95.0 \\
        CaCoVID (AV ext.) & 20\% & 54.4 & 49.7 & 49.7 & 77.1 & 95.6 \\
        \textbf{\method} & 20\% & \textbf{55.3} & \textbf{50.7} & \textbf{50.5} & \textbf{78.3} & \textbf{97.1} \\
        \midrule
        Random & 10\% & 24.5 & 22.3 & 22.2 & 34.6 & 42.9 \\
        FastV (AV ext.) & 10\% & 45.3 & 41.7 & 41.4 & 64.4 & 79.9 \\
        ToMe (AV ext.) & 10\% & 47.5 & 43.4 & 43.4 & 67.0 & 83.4 \\
        FastAV & 10\% & 51.7 & 47.6 & 47.2 & 73.5 & 91.2 \\
        CaCoVID (AV ext.) & 10\% & 52.7 & 48.5 & 48.1 & 74.9 & 92.9 \\
        \textbf{\method} & 10\% & \textbf{53.3} & \textbf{48.9} & \textbf{48.7} & \textbf{75.4} & \textbf{93.6} \\
        \bottomrule
    \end{tabular}%
    }
\end{table}

\begin{figure}[!t]
    \centering
    \includegraphics[width=1.0\textwidth]{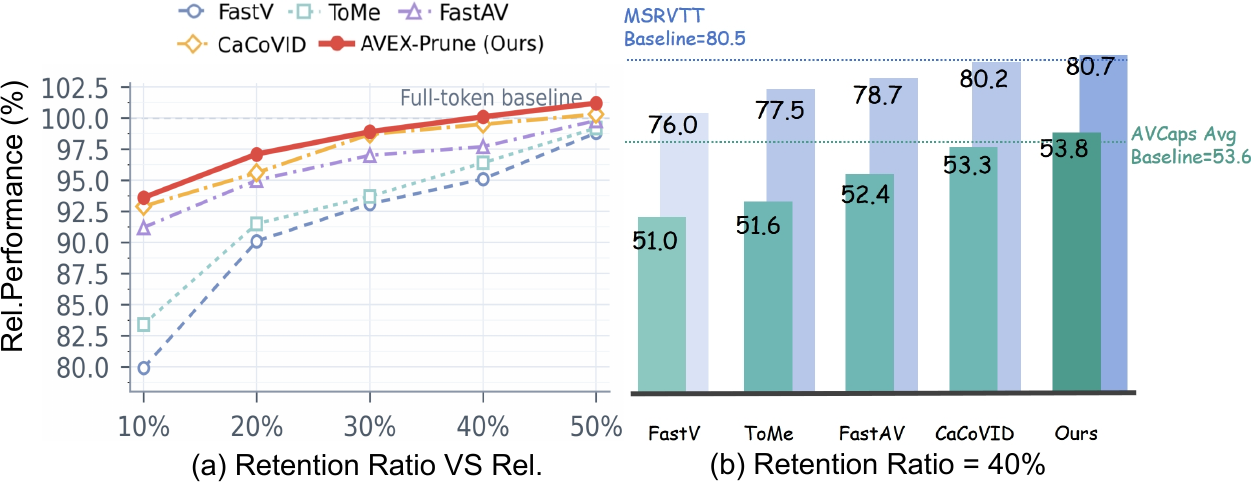}
    \caption{Performance under audio-visual token pruning. Left: relative performance across
    retention ratios. Right: AVCaps $C_{av}$, $C_v$, $C_a$ at a 40\% retention ratio.}
    \label{fig:performance}
\end{figure}

Table~\ref{tab:main_pruning} shows that \method{} consistently outperforms all baselines,
with the largest margin under low retention ratios. The learned ranking transfers to MSRVTT
without audio conditioning. Figure~\ref{fig:performance} summarises the full sweep.

\subsection{Ablation Studies}
\label{sec:ablation}

We then isolate the training signal. Table~\ref{tab:ablation} compares the full four-way
exchange objective with warmup-only training, random equal-cost exchange, within-modality
exchange, and variants that remove one within-modality branch. All rows use the same 20\%
retention ratio and the same number of reward evaluations when applicable.

\begin{table}[t!]
    \centering
    \scriptsize
    \setlength{\tabcolsep}{1.6pt}
    \renewcommand{\arraystretch}{0.82}
    \caption{Training-signal ablation and group size (20\% retention ratio).}
    \label{tab:ablation}
    \resizebox{0.78\textwidth}{!}{%
    \begin{tabular}{lccccccc}
        \toprule
        \textbf{Variant} & \textbf{V-V} & \textbf{A-A} & \textbf{V-A} & \textbf{A-V} & \textbf{$C_{av}$} & \textbf{$C_v$} & \textbf{$C_a$} \\
        \midrule
        Warmup only & \xmark & \xmark & \xmark & \xmark & 51.8 & 47.5 & 41.5 \\
        Random exchange & \xmark & \xmark & Mixed & Mixed & 53.6 & 48.9 & 45.7 \\
        V-to-V only & \cmark & \xmark & \xmark & \xmark & 53.4 & 50.2 & 43.0 \\
        A-to-A only & \xmark & \cmark & \xmark & \xmark & 52.8 & 46.0 & 47.2 \\
        V-to-V + A-to-A & \cmark & \cmark & \xmark & \xmark & 54.7 & 49.8 & 47.9 \\
        w/o A-to-A & \cmark & \xmark & \cmark & \cmark & 54.9 & 50.5 & 45.0 \\
        w/o V-to-V & \xmark & \cmark & \cmark & \cmark & 54.6 & 48.0 & 49.9 \\
        \textbf{Full \method} & \cmark & \cmark & \cmark & \cmark & \textbf{55.3} & \textbf{50.7} & \textbf{50.5} \\
        \midrule
        \multicolumn{8}{c}{\textit{Group size ablation ($g$)}}\\
        \midrule
        $g=4$ & \cmark & \cmark & \cmark & \cmark & 54.9 & 50.3 & 50.1 \\
        $g=8$ & \cmark & \cmark & \cmark & \cmark & \textbf{55.3} & \textbf{50.7} & \textbf{50.5} \\
        $g=16$ & \cmark & \cmark & \cmark & \cmark & 55.0 & 50.5 & 50.3 \\
        $g=32$ & \cmark & \cmark & \cmark & \cmark & 54.6 & 49.9 & 49.7 \\
        \bottomrule
    \end{tabular}%
    }
\end{table}

Warmup alone uses a static attention-based ranking incapable of per-sample adaptation.
Adding exchange supervision substantially improves all metrics; removing A-to-A reduces
$C_a$ and removing V-to-V lowers $C_v$ and $C_{av}$, confirming that within-modality ranking
and cross-modal competition are complementary. All rows use a 20\% retention ratio.

\begin{figure}[t!]
    \centering
    \includegraphics[width=1.0\textwidth]{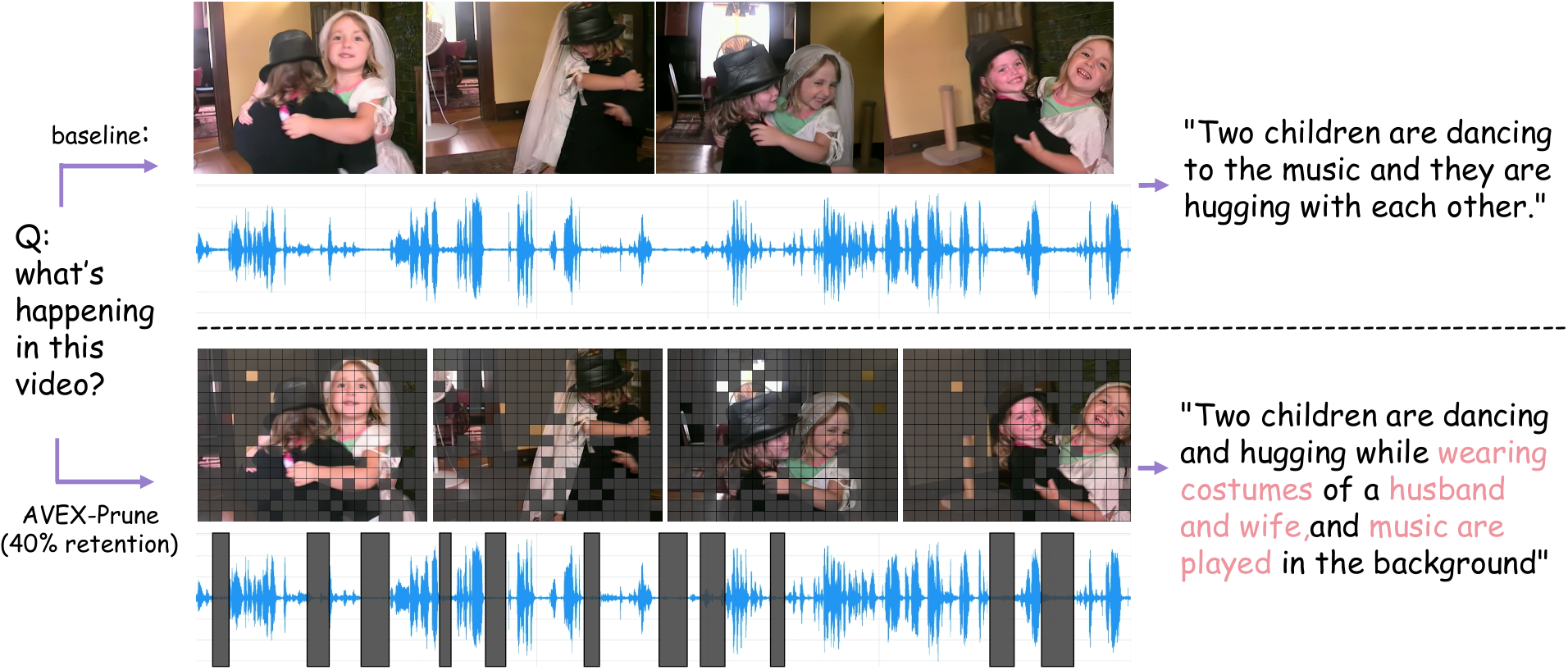}
    \caption{Qualitative captioning at a 40\% retention ratio. \method{} preserves visual and
    acoustic details omitted by baselines.}
    \label{fig:qual_caption}
\end{figure}

\subsection{Efficiency and Reproducibility}
\label{sec:efficiency}

Exchange branches are training-only. At inference, \method{} runs one policy pass, keeps
the selected tokens, and invokes the captioning backbone once. Table~\ref{tab:efficiency}
reports cost at a 40\% retention ratio on one A800 GPU with batch 1. FLOPs measure multimodal
prefill and LLM forward computation; prefill latency measures context construction; end-to-end
latency includes token scoring, selection, and caption decoding; memory is peak allocated
GPU memory during inference.

\begin{table}[t!]
    \centering
    \scriptsize
    \setlength{\tabcolsep}{2.4pt}
    \renewcommand{\arraystretch}{0.92}
    \caption{Inference efficiency at a 40\% retention ratio (1$\times$A800, batch 1).}
    \label{tab:efficiency}
    \begin{tabular}{lcccc}
        \toprule
        \textbf{Method} & \textbf{FLOPs (T)} & \textbf{Prefill (ms)} & \textbf{E2E (ms)} & \textbf{Mem. (GB)} \\
        \midrule
        Full tokens & 14.8 & 412 & 658 & 22.4 \\
        FastV (AV ext.) & 6.1 & 178 & 376 & 14.2 \\
        ToMe (AV ext.) & 6.0 & 174 & 371 & 14.0 \\
        FastAV & 6.2 & 181 & 379 & 14.4 \\
        \textbf{\method} & 6.3 & 188 & 386 & 14.9 \\
        \bottomrule
    \end{tabular}
\end{table}

\method{} reduces FLOPs by over 55\% and prefill latency by over 50\% compared with
full-token inference. The policy adds only 10 to 15~ms end-to-end overhead since reward
and exchange branches are training-only. Training uses four A800 GPUs for 18 GPU hours;
only the 30M policy is updated.

\section{Conclusion}
\label{sec:conclusion}

In this paper, we investigate audio-visual token pruning for efficient captioning,
proposing that audio and visual tokens should be ranked through their measured interaction,
not as independent pools. Our exchange strategy replaces low-confidence retained tokens with
high-confidence candidates from the same or the other modality, using CIDEr differences from
these swaps as direct supervision to learn which tokens truly contribute to caption quality.
Results on AVCaps confirm that this exchange-based ranking preserves full-token caption
quality at a 40\% retention ratio on both VILA~1.5-8B and VideoLLaMA~2, with ablation
studies showing that all four exchange types provide complementary supervision signals.
Building on these insights, we introduce \method, an RL-based dynamic token pruning method
that learns token selection from actual caption reward improvement, reducing FLOPs by over
55\% with only 10 to 15~ms overhead at inference. This framework not only validates the
effectiveness of exchange-based cross-modal supervision but also offers a general approach
for dynamic budget allocation across modalities with very different properties.

\clearpage
\fontsize{8pt}{9.5pt}\selectfont

\end{document}